\documentclass[11pt,a4paper]{article}
\usepackage[hyperref]{acl2021}
\usepackage{times}
\usepackage{latexsym}

\usepackage{microtype}

\usepackage{amssymb}
\usepackage{bm}
\usepackage{times}

\usepackage{graphicx}
\usepackage{multirow}
\usepackage{enumerate}
\usepackage{booktabs}
\usepackage{enumitem}
\usepackage{amsmath}
\usepackage{hyperref}

\aclfinalcopy  

\title{GeoQA: A Geometric Question Answering Benchmark Towards Multimodal Numerical Reasoning}

\author{Jiaqi Chen$^{1}$\thanks{~~Equal contribution.}~,~~ Jianheng Tang$^{2\ast}$,~~ Jinghui Qin$^{1}$,~~ Xiaodan Liang$^{2}$\thanks{~~Corresponding author.}~, \\ 
{\bf Lingbo Liu}$^{1,3}${\bf ,}~~{\bf Eric P. Xing}$^{4}${\bf ,}~~ {\bf Liang Lin}$^{1,3}$ \\
{\small $^1$Sun Yat-sen University, $^2$Shenzhen Campus of Sun Yat-sen University, $^3$Dark Matter AI Inc., } \\
{\small $^4$Mohamed bin Zayed University of Artificial Intelligence}\\
{\small\texttt{\{jadgechen,sqrt3tjh,xdliang328,liulingbo918\}@gmail.com, }}\\
{\small\texttt{qinjingh@mail2.sysu.edu.cn, eric.xing@petuum.com, linliang@ieee.org}}
}

\begin{document}
\maketitle
\begin{abstract}

Automatic math problem solving has recently attracted increasing attention as a long-standing AI benchmark. In this paper, we focus on solving geometric problems, which requires a comprehensive understanding of textual descriptions, visual diagrams, and theorem knowledge. However, the existing methods were highly dependent on handcraft rules and were merely evaluated on small-scale datasets.
Therefore, we propose a \textbf{Geo}metric \textbf{Q}uestion \textbf{A}nswering dataset \textbf{GeoQA}, containing 4,998 geometric problems with corresponding annotated programs, which illustrate the solving process of the given problems. Compared with another publicly available dataset GeoS, GeoQA is 25 times larger, in which the program annotations can provide a practical testbed for future research on explicit and explainable numerical reasoning.
Moreover, we introduce a Neural Geometric Solver (NGS) to address geometric problems by comprehensively parsing multimodal information and generating interpretable programs. We further add multiple self-supervised auxiliary tasks on NGS to enhance cross-modal semantic representation.
Extensive experiments on GeoQA validate the effectiveness of our proposed NGS and auxiliary tasks. However, the results are still significantly lower than human performance, which leaves large room for future research.
Our benchmark and code are released at \href{https://github.com/chen-judge/GeoQA}{https://github.com/chen-judge/GeoQA}.

\end{abstract}

\section{Introduction}



In recent years, developing machine learning systems to solve math word problems (MWPs) automatically has attracted increasing attention due to its high academic value and the great application potential in smart education~\cite{bajaj2018smart,lin2018intelligent}. Most of the existing methods focus on solving arithmetic and algebraic problems, including traditional machine learning approaches~\cite{kushman2014learning, zhou-etal-2015-learn, huang-etal-2016-well} and network-based models~\cite{dns, seq2et, seq2tree}, while solving geometric problems has been rarely investigated ~\cite{seo2014diagram,seo2015solving,sachan2017textbooks}. 
As a classic math problem, geometry dominates a large portion of  secondary education. Due to its challenges and data characteristics, geometry problem can also serve as a multimodal numerical reasoning benchmark requiring joint reasoning over diagram and text.




\begin{figure}[t]
\begin{center}
 \includegraphics[width=0.95\columnwidth]{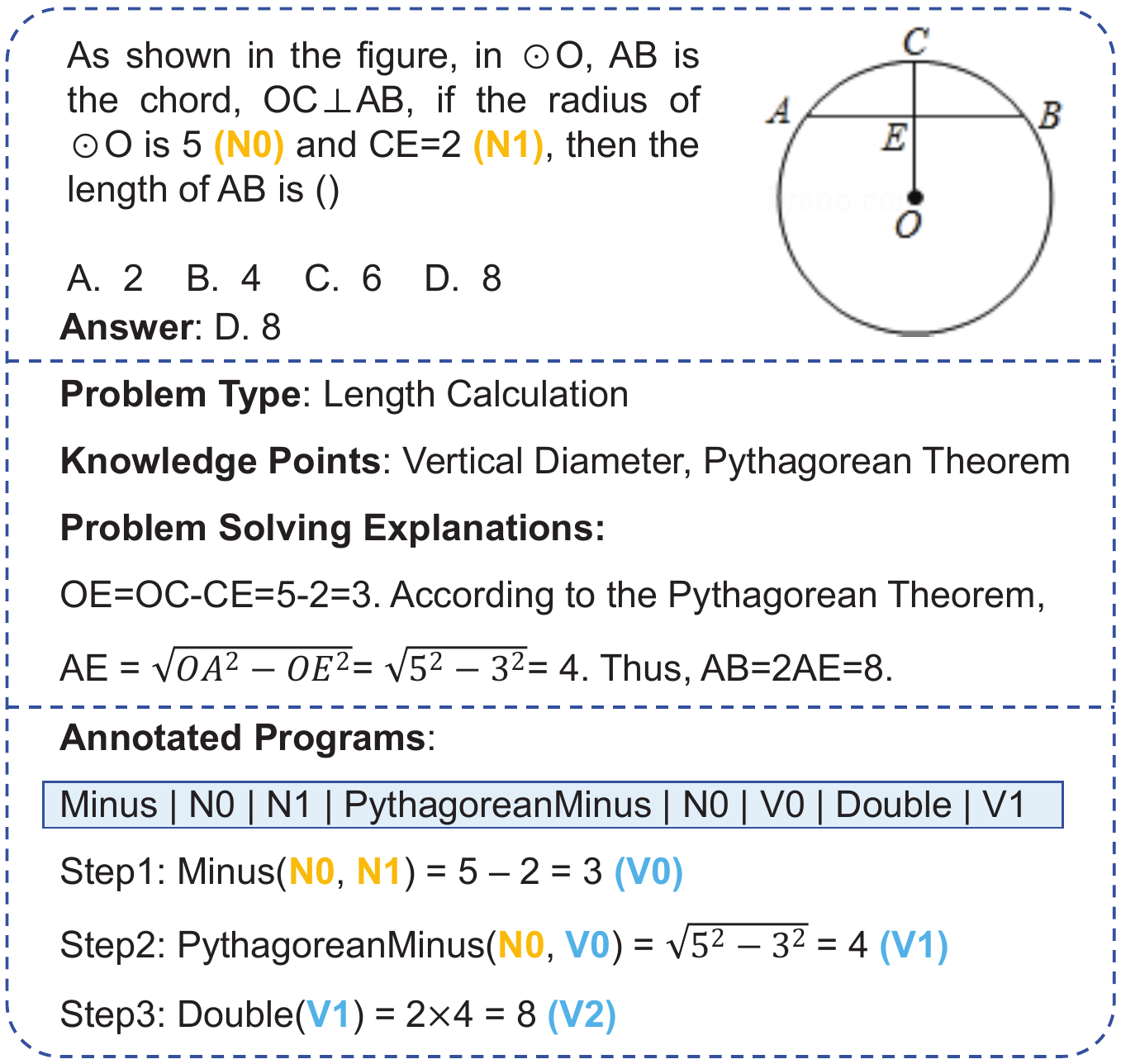}
\end{center}
  \caption{Illustration of a typical geometry problem with the annotated programs in our GeoQA dataset. 
  }
\label{fig:introduce}
\end{figure}

As shown in Figure~\ref{fig:introduce}, a typical geometric question mainly consists of textual descriptions and geometric diagrams.
Compared with math word problems, which only involve problem texts, geometric questions have posed the following new challenges. {\bf{First}}, the additional problem diagrams provide essential information absent from the problem text, such as the relative location of lines and points; thus, the solver should have the capability to parse the diagram. {\bf{Second}}, to solve a geometry problem, we need to understand and align the semantics of text and diagram simultaneously.  However, the problem text often includes some ambiguous references and implicit relations to diagram elements, which increases the difficulty of joint reasoning over text and diagram. {\bf{Third}}, many geometric problems require extra theorem knowledge in the problem solving process. For example, in Figure~\ref{fig:introduce}, the Pythagorean Theorem is used to calculate the length of line $AE$.

Though some previous methods \cite{seo2014diagram,seo2015solving,sachan2017textbooks,sachan2020discourse,sachan2020knowledge} attempt to resolve the mentioned issues, the performance of their geometric problem solving systems is far away from satisfactory. They highly depended on limited handcraft rules and were only validated on small-scale datasets, making it hard to generalize to more complex and real-world cases. Besides, the solving process is sophisticated, which means it is difficult for a human to understand and examine its reliability.

To refresh the research on geometric problem solving and promote further study on multimodal numerical reasoning, we propose a large-scale real-world geometric question answering dataset called GeoQA, which contains 4,998 multiple-choice geometric problems collected from real math exams in Chinese middle school. Inspired by~\citet{amini2019mathqa}, we additionally introduce a new domain-specific language to model precise operation programs corresponding to the geometry problem. These executable programs represent the numerical reasoning steps of geometry problems. 
Compared with the existing dataset GeoS and GeoS++~\cite{seo2015solving,sachan2017textbooks}, 
our GeoQA is larger, more diverse, provides additional program annotation, thus serves as a promising benchmark to improve both generalization and interpretability of the multimodal numerical reasoning approaches.


Moreover, we propose the first deep learning-based approach for geometry problem solving, named as Neural Geometric Solver (NGS). It applies a co-attention mechanism to fuse the representation of text and diagram, and predicts the explainable programs based on the cross-modal representation. These sequential programs can be executed to obtain a final answer.  Benefiting from the structured and explainable program prediction, our NGS has both merits in superior performance by learning-based models compared to previous rule-based methods, as well as generating explainable numerical reasoning steps via the program sequence in favor of the model diagnosis. We further design three highly-relevant pretext tasks to enhance text-diagram semantic representation, including diagram jigsaw location prediction, diagram geometric element prediction, and knowledge point prediction. Extensive experiments are conducted on GeoQA benchmark, and the quantitative comparisons show the superiority of the proposed NGS and auxiliary tasks over other multimodal baselines.





In summary, our contributions are three-fold:
\begin{itemize}[leftmargin=*]
\setlength{\itemsep}{0pt}
\setlength{\parsep}{0pt}
\setlength{\parskip}{0pt}
\item We construct a large-scale dataset for geometric problem solving, which contains 4,998 Chinese geometric multiple-choice questions with rich domain-specific program annotations. 
\item A novel Neural Geometric Solver is proposed to solve geometric problems by generating symbolic programs based on the joint understanding of textual descriptions and diagrams. 
\item Multiple specialized auxiliary tasks are employed to effectively improve the semantic representation of text and diagrams. Experiments show the superiority of our NGS equipped with these auxiliary tasks.
\end{itemize}

\section{Related Work}
\paragraph{Geometry Problems Solving} 
Developing automated systems to solve geometry problems has a long history in AI~\cite{old1, old2,old3,jgex}. 
For example, \citet{seo2014diagram,seo2015solving} built the first automated system, GeoS, to solve SAT style geometry problems. GeoS used NLP and computer vision techniques (e.g., OCR) to parse a geometry problem’s text and diagram jointly as logic forms.
However, this system highly depended on the manually designed logic forms and was only examined in a small dataset with 185 problems. Besides, the limited logic forms are hard to cover various geometry problems, leading to low generalization. To improve GeoS,   \citet{sachan2017textbooks, sachan2017learning} replaced these handcraft constraints with geometry axiomatic knowledge in the form of horn-clause rules, but their dataset and code are not released. 
To boost the generalization and interpretability of existing works, we propose a large-scale GeoQA benchmark, which is 25 times larger than the only public dataset \citet{seo2015solving}, and provides program annotation. 
 



\paragraph{Multimodal Reasoning}
Visual question answering is a representative multimodal task that requires the model to have reasoning ability~\cite{goyal2017making,yu2019deep}. ~\citet{johnson2017clevr} built a new diagnostic VQA dataset (called CLEVR) with annotated functional programs. Based on this benchmark, some methods proposed an implicit reasoning framework to jointly encode multimodal information~\cite{perez2017film,santoro2017simple}. Moreover, several works~\cite{yi2018neural,mao2019neuro} utilize domain-specific languages to perform explicit symbolic reasoning. However, these program languages only consider elementary operations, such as counting objects. They are not directly applicable to geometric problems, which require multiple steps of numerical calculation and involve theorem knowledge. 


\paragraph{Self-supervised Auxiliary Task}
Self-supervised pretraining has gradually emerged~\cite{doersch2017multi,newell2020useful} as a effective technique to deal with label scarcity and improve model performance. To enhance visual features, most of these methods construct pseudo labels automatically and train on auxiliary tasks, including image jigsaw \cite{noroozi2016unsupervised,ahsan2019video}, inpainting~\cite{pathak2016context}, super resolution~\cite{ledig2017photo}, etc. Inspired by these works, we design two self-supervised auxiliary tasks and a supervised auxiliary task to enhance the reasoning ability of our NGS.



\section{GeoQA Dataset}
Due to the limited data scale and problem types, the existing geometric problem reasoning dataset~\cite{seo2015solving} can neither comprehensively reflect model's reasoning ability, nor support the training of neural models.
To propose a better benchmark for the evaluation of multimodal numerical reasoning and inspire applications in smart education, we collect a new dataset GeoQA. It contains 4,998 diverse real-world geometric problems in Chinese middle school exams, and each problem is additionally annotated by specific programs that describe the problem solving process. Besides, we also provide human performance on GeoQA, as shown in Table~\ref{table-baselines}. 



\begin{table}[tbp]
\centering
\resizebox{1.0\linewidth}{!} {
\begin{tabular}{c|c|c|c|c}
\hline
& Total & Train & Val & Test \\
\hline
\hline
Number & 4998 & 3499 & 745 & 754 \\
\hline
Angle & 2737 & 1932 & 388 & 417\\
Length & 1869 & 1300 & 286 & 283\\
Other & 392 & 267 & 71 & 54\\
\hline
\#Avg DS & 108$\times$140 & 108$\times$140 & 107$\times$141 & 107$\times$140\\
\#Avg QL & 52.5 & 52.4 & 52.4 & 52.7\\
\#Avg KP & 2.10 & 2.10 & 2.07 & 2.14\\
\#Avg ET & 1.11 & 1.13 & 1.08 & 1.09\\
\hline
\#Avg OP & 1.98 & 1.99 & 1.92 & 1.98\\
\#Avg PL & 5.35 & 5.39 & 5.17 & 5.36\\
\hline
\end{tabular}
}
\caption{Statistics of our GeoQA dataset. It contains three types of problems, including angle, length, and other problems. Besides, DS, QL, KP, and ET represent diagram size, question length, knowledge points, and element types, respectively. OP and PL represent operation number and program length.
}
\label{table-dataset}
\end{table}

\subsection{Data Description}
 Generally, a geometry multiple-choice problem can be represented as a tuple ($t$, $d$, $c$, $i$) where $t$ is the problem text in natural language, $d$ is the problem diagram, $\mathbf{c}=(c_1,c_2,c_3,c_4)$ represents the 4 numerical options for the problem, $i$ is the answer index. Given the text $t$ and diagram $d$, an algorithm is required to predict the correct answer $c_i \in \mathbf{c}$. To collect as much useful information as possible, we also provide the natural language-based problem solving explanations $e$, problem type $t$, the related knowledge points $k$, and our annotated programs $p$ for each problem. Therefore, a geometry problem can be represented as ($t$, $d$, $c$, $i$, $e$, $t$, $k$, $p$). 
 Fig. \ref{fig:introduce} shows an example of geometric problems.

Moreover, there are three problem types in our GeoQA, i.e., angle calculation, length calculation, and others which contain various types of problems such as area calculation. 
We adopt the corpus diversity metric proposed by \citet{miao2020diverse} to evaluate the diversity of GeoQA. The result is 0.47, which is relatively high compared with other math problem datasets, indicating that our dataset is diverse. 
The source data already contains manually tagged knowledge points in each problem, we design rule-based regular expressions to normalize the original knowledge points to 50 categories.
We split our GeoQA into three subsets -- train set, valid set, and test set, in a ratio of 7.0: 1.5: 1.5. 
The data statistics of our GeoQA are shown in Table~\ref{table-dataset}.

\subsection{Program Representation}
The neural network has proved to be a powerful tool to address complex multimodal reasoning tasks. However, it still faces challenges when conducting numerical calculation and providing explicit problem solving process, which are actually two crucial points in the task of geometric problem solving. To make better use of neural networks in the geometric problems solving process, inspired by~\citet{amini2019mathqa}, we introduce a new domain-specific language to model the geometric problem solving process based on the GeoQA dataset. This program language can be directly executed to calculate the numerical answer based on the predefined operations and their arguments.

The program types contain operations $OP$, constants $Const$, problem variables $N$, and process variables $V$. As shown in Table~\ref{table-program}, operations are divided into multiple categories, including Basic, Arithmetic, Trigonometric, Theorem, and Formula operations. Each operator involves $n (=1, 2, 3)$ elements selected from constants and variables. Constants are predefined numbers that are frequently used in geometric problems, such as $\pi$ and the degree of a Right Angle (90). The problem variables refer to all the number that appears in the current problem, and process variables are obtained during the execution process.

In addition to the common math operations, our programs also contain some operations representing the knowledge of theorems and formulas that is helpful to address geometric problems, such as the Pythagorean theorem and the area calculation formula of a circle. It is worth noting that many simple geometric formulas do not require additional definitions. For example, given a square with side length $a$, its area can be directly computed by $\text{Multiply}(a,a)$. 

The interpretability of our program is reflected on the sequential process of the operations, the selected constants and variables, and the application of theorems and formulas. As shown in Figure~\ref{fig:introduce}, we can have a general understanding of the entire problem solving process after reading the program.







\begin{table}[tbp]
\centering
\resizebox{1.0\linewidth}{!}{
\begin{tabular}{c|c}
\hline
Types & Programs \\
\hline
\hline
Basic & Equal, Double, Half \\
\hline
Arithmetic & Add, Minus, Multiply, Divide\\
\hline
Trigonometric & Sin, Cos, Tan, Arc-Sin, Arc-Cos\\
\hline
Theorem & Pythagorean Add/Minus, Proportion, \\
\& Formula & Circle Area, Circle Perimeter, Cone Area\\
\hline
Constant & 30, 60, 90, 180, 360, $\pi$, 0.618\\
\hline
\end{tabular}
}
\caption{An overview of 18 operations of four different types and 7 constants in the defined program set.
}
\label{table-program}
\end{table}

\subsection{Collection and Annotation}
We collect GeoQA from two online education websites\footnote{\href{http://www.zxxk.com/}{http://www.zxxk.com/} and   \href{http://www.jyeoo.com/}{http://www.jyeoo.com/}}. These problems are oriented grades 6-12, containing various types of problems with corresponding knowledge points and solving explanations. 
we organize more than ten well-trained college students with a relevant major to specifically annotate our programs by referring to the solving explanations. To ensure label quality and consistency, they are required to read the guideline of annotation standards and examples in advance. Each annotated program is double-checked by one of the authors, and the annotator with low accuracy would be disqualified. 
The annotated operations required to solve the problem are limited to a maximum of 4 steps, thus a small number of complex and hard problems are filtered.






\section{Neural Geometric Solver}
\begin{figure*}[t]
\begin{center}
 \includegraphics[width=0.97\textwidth]{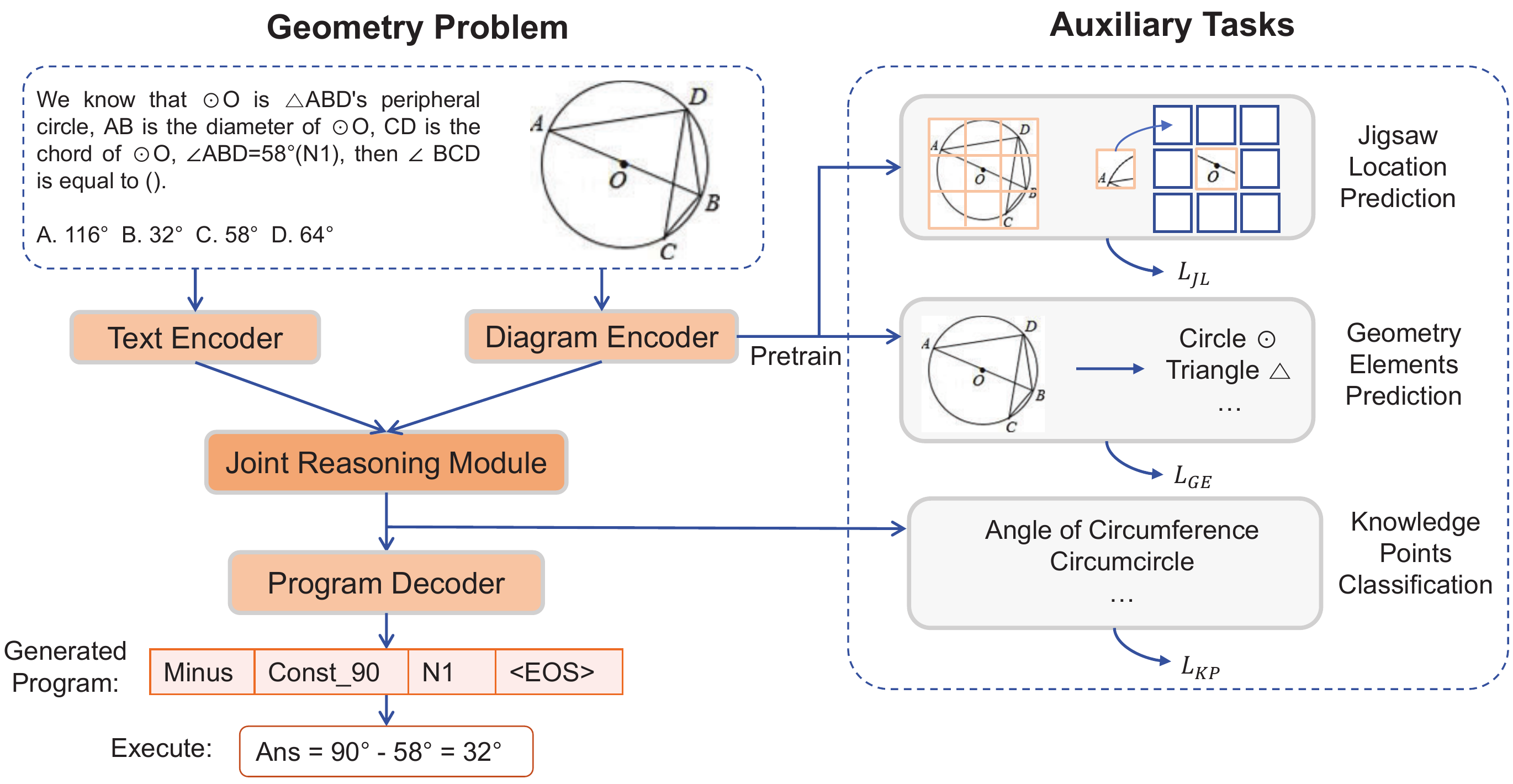}
\end{center}
  \caption{The overall architecture of our Neural Geometric Solver (left) in conjunction with auxiliary tasks (right). The problem text and diagram is encoded separately, then fed into a joint reasoning module together to obtain cross-modal fusion of text and diagram features. A decoder utilizes fused multimodal features to generate the interpretable programs. In addition, we propose three auxiliary tasks to enhance feature representation and facilitate multimodal reasoning.
}
\label{fig:achitecture}
\end{figure*}

We propose Neural Geometric Solver (NGS) to address geometric problems by jointly understanding text, diagram, and then generating explainable programs. Moreover, we utilize some novel auxiliary tasks to enhance the understanding ability of our NGS.
The overall architecture of our NGS is shown in Fig.~\ref{fig:achitecture}. 


\subsection{The Architecture of NGS}

\subsubsection{Problem Encoder}
\paragraph{Text Encoder}
Given a problem text $P$ = $\left \{ x_i \right \}^n_{i=1}$, each token $x_i$ is first embedded into a word vector $\mathbf{x}_i$. A single-layer unidirectional LSTM~\cite{LSTM} is then applied to encode each word embedding $\mathbf{x}_i$ into a hidden state $\mathbf{h}_i$. The sequence of the hidden state in LSTM are used to represent problem text $P$ as $H_P= [\mathbf{h}_0;...;\mathbf{h}_n]$.


\paragraph{Diagram Encoder}
For representing a problem diagram, we apply the first three stages of a ResNet-101~\cite{he2016deep}  to extract it as the diagram feature, which can be formalized as a feature matrix $H_D$ $\in$ $\mathbb{R}^{ m \times d }$.
Moreover, we also apply multiple auxiliary tasks for pretraining the diagram encoder.
Note that the parameters of the diagram encoder are fixed when training the overall NGS.

\subsubsection{Joint Reasoning Module}
Given the text feature $H_P$ and the diagram feature $H_D$, it is crucial for solving geometric problems to jointly fuse and align the cross-modal information.
To this end, inspired by~\citet{yu2019deep}, we adopt a co-attention module to conduct cross-modal joint reasoning with attention mechanism.  
The co-attention module consists of 12 self-attention (SA) units and 6 guided-attention (GA) units, which fully fuse and align the text representation $H_P$ and the diagram representation $H_D$. 
$H_P$ is first encoded by 6 self-attention units (i.e., original Transformer), and the final hidden state processed by the 6-th self-attention unit is used as guiding information. 
Then, the guiding information is fed into another stacked 6 self-attention units and 6 guided-attention units to achieve cross-modal semantic fusion and alignment.
Finally, the co-attention module outputs a cross-modal representation $F_D=[f^D_1;...;f^D_n]$, which contains rich information over the problem text and diagram. 

In this work, we find that text information is more fundamental than diagram information. Therefore, we further enhance the cross-modal representation with the help of textual information. Specifically, we concatenate $H_P$ and $F_D$ to acquire an enhanced reasoning module output $F_R$ for decoding programs.

Besides, an attentional reduction network with a two-layer MLP is applied to aggregate feature $F_D$ 
into $\tilde{F}_D$.
Similarly, we concatenate $\tilde{F}_D$ and the last encoder state of the text encoder $\bm h_n$, obtaining $\tilde{F}_R$ as the final gathered multimodal feature vector.

\subsubsection{Program Decoder}

The program decoder module generates the programs sequentially under the guidance of multimodal information. Concretely, we use a LSTM decoder~\cite{LSTM} with attention~\cite{bahdanau2014neural} over the reasoning module output $F_R$ . Let $\{y_t\} (1\leq t \leq T)$ be the target program to be generated and $\bm s_t$ be the hidden state of LSTM at time step $t$. $\tilde{F}_R$ is fed into a linear layer to obtain the initial state $\bm s_0$. $\bm s_t$ is concatenated with the attention result and fed to a linear layer with the softmax function to predict the distribution of the next program token $P_t$.

During training, the generation loss $L_g$ is the negative log-likelihood (NLL) of the target program:
{
\setlength\abovedisplayskip{0.45pt}
\setlength\belowdisplayskip{0.55pt}
\begin{equation*}
    \mathcal L_g(\bm \theta)=\frac{1}{T}\sum_{t=1}^T \log P_t(y_t|\bm x, y_1, ..., y_{t-1};\bm \theta),
\end{equation*}
}%
where $\bm \theta$ are the parameters of the entire NGS architecture except for the diagram encoder, $\bm x$ is the input of both problem text and the extracted diagram feature. When testing, the decoder only observes the input text and diagram feature along with the program parts that have been generated.

\subsubsection{Program Executor}

After a beam of top $N$ program sequences $\{g_1,...,g_n\}$ are generated from the program decoder, the executor computes them step by step. When executing the program, the token sequence is first divided into several parts based on the position of operators in the program. Once a complete operation program has been decoded, each operator in the program is executed sequentially to obtain a numerical result. The execution process fails if $g_i$ has a grammar error (e.g., the number of augments does not match the current operator), or the executed value does not match any options in the current problem. NGS adopts the first successfully executed program as the predicted solution and chooses the corresponding operation. If all $N$ program sequences fail, the executor will report ``no result'' directly instead of guessing an option. Fig.~\ref{fig:introduce} shows the detailed step-by-step execution of a final predicted program sequence.






\subsection{Auxiliary Tasks}
\subsubsection{Self-supervised Diagram Auxiliary Task}

Although our NGS can jointly fuse text feature and diagram feature with a co-attention mechanism, a powerful diagram encoder is needed to improve the problem understanding and answer accuracy. To obtain a high-quality diagram feature, we investigate two self-supervised auxiliary tasks, named as \textbf{Jigsaw Location Prediction} and \textbf{Geometry Elements Prediction}, to pretrain diagram encoder. 

\paragraph{Jigsaw Location Prediction}
In the Jigsaw location prediction task that enforces pixel-level perception, we first split a diagram as $m \times m$ blocks and select the center block as the target. Then, we shuffle other blocks randomly and train the diagram encoder to predict the correct relative location between these shuffled blocks and the target using a cross-entropy loss. 

\paragraph{Geometry Elements Prediction}
For object-level understanding, we design a geometry elements prediction task that aims at training the diagram encoder to predict the geometry elements appearing in the diagram. A diagram usually contains multiple geometry elements which are also mentioned in the problem text and the solving explanation. We extract these geometry elements from text as the label and deploy an N-way classifier with binary cross-entropy (BCE) as the loss function to train the diagram encoder, where $N$ is the number of the possible geometry elements on the training set. 

\subsubsection{Knowledge Points Prediction}
In addition to self-supervised diagram training, we also propose another auxiliary learning task called knowledge points prediction to enhance a problem's overall representation by providing an extra training signal. We summarize about 50 knowledge points for our GeoQA, and label each problem with one or more knowledge points. We predict the knowledge points for each problem based on the gathered feature vector $\tilde{F}_R$ outputted from the joint reasoning module. Different from the diagram training, the knowledge points prediction task is trained with NGS simultaneously. 
We also deploy a K-way classifier with binary cross-entropy (BCE) as the loss function to train the knowledge points prediction multi-label task, where $K$ is the total number of the possible knowledge points on the training set.

\begin{table*}[htbp]
\centering
\resizebox{0.9\linewidth}{!} {
\begin{tabular}{c|c|c|c|c|c}
\hline
\multicolumn{2}{c|}{Method} & Total (\%) & Angle (\%) & Length (\%) & Other (\%)  \\
\hline\hline
\multirow{2}{*}{Human} 
& Text-Only & 63.0 & 58.0 & 71.7 & 55.6 \\
& Text-Diagram & 92.3 & 94.2 & 90.5 & 87.0 \\
\hline
\multirow{3}{*}{W/O Program} 
& FiLM \cite{perez2017film} & 32.8 & 33.6 & 32.9 & 25.9 \\
& RN \cite{santoro2017simple} & 38.2 & 42.2 & 34.3 & 27.8 \\
& MCAN \cite{yu2019deep} & 39.5 & 43.2 & 36.0 & 29.6 \\

\hline

\multirow{2}{*}{Text-Only}
& Seq2Prog \cite{amini2019mathqa}& 54.2 & 66.4 & 38.5 & 42.6 \\
& BERT2Prog \cite{devlin2018bert}& 54.4 & 65.7 & 41.0 & 37.0 \\

\hline

\multirow{4}{*}{Text-Diagram} 
& BERT2Prog + Diagram & 52.5 & 65.7 & 37.1 & 31.5 \\
& Seq2Prog + Diagram & 53.4 & 62.4 & 44.5 & 31.5 \\
& NGS (Ours) & 57.4 & 68.6 & 43.5 & \textbf{44.4} \\
& NGS-Auxiliary (Ours) & \textbf{60.0} & \textbf{71.5} & \textbf{48.8} & 29.6 \\
\hline
\end{tabular}
}
\caption{The answer accuracy comparison on different test subsets of GeoQA dataset. ``Human", ``W/O Program", ``Text-Only", and ``Text-Diagram" refer to the performance of human, not using the program, using text modal only, and conducting multimodal numerical reasoning on both text-diagram modals, respectively.}
\label{table-baselines}
\end{table*}

\section{Experiments}

\begin{table}[htbp]
\centering
\resizebox{0.9\linewidth}{!} {
\begin{tabular}{c|c|c|c}
\hline
Method & BS & Acc(\%) & NR(\%) \\
\hline
\hline
 & 1 & 33.8 & 54.0 \\
 Seq2Prog + Diagram & 10 & 53.4 & 19.5 \\
& 100 & 59.9 & 3.91 \\
\hline
& 1 & 46.3 & 41.2 \\
NGS-Auxiliary & 10 & 60.0 & 13.5 \\
& 100 & 63.9 & 2.86 \\
\hline
\end{tabular}
}
\caption{Performance comparison under different beam size settings. BS, Acc, and NR represent beam size, accuracy, and no result, respectively.}
\label{table-beam}
\end{table}

\subsection{Experimental Setup and Training Details }

We conduct experiments on GeoQA dataset, and adopt answer accuracy as the evaluation metric. Although there is another  available geometric problem dataset~\cite{seo2015solving}, the limited data scale (with only 67 training samples) makes it impossible to support neural network training. On the other hand, previous geometry problem solving systems require additional inputs (e.g.,  OCR and dependency parsing results) of the problem ~\cite{jgex, seo-etal-2015-solving} or not release their codes~\cite{sachan2017learning,sachan2020knowledge}. Therefore, they are not comparable on our GeoQA dataset.




\textbf{Implementation Details:}
In this work, we implement the proposed method with Pytorch \cite{paszke2017automatic}. 
The learning rate is $1e^{-3}$ and the batch size is set to 32. All models are trained around 100 epochs and optimized by Adam optimizer \cite{kingma2014adam}. The beam size is typically set to 10.
When pretraining the diagram encoder, 
we first fill the diagram with a white background to make it equal in length and width, and resize it to $224 \times 224$.
Then, we utilize the diagram feature extracted by the encoder to predict jigsaw location and geometry elements simultaneously and optimize the diagram encoder to obtain an informative diagram feature with a learning rate of $1e^{-5}$. Finally, the loss weight of the knowledge points classification task is set to $1$, to promote the overall understanding of problems.

\begin{figure*}[t]
\begin{center}
 \includegraphics[width=0.99\textwidth]{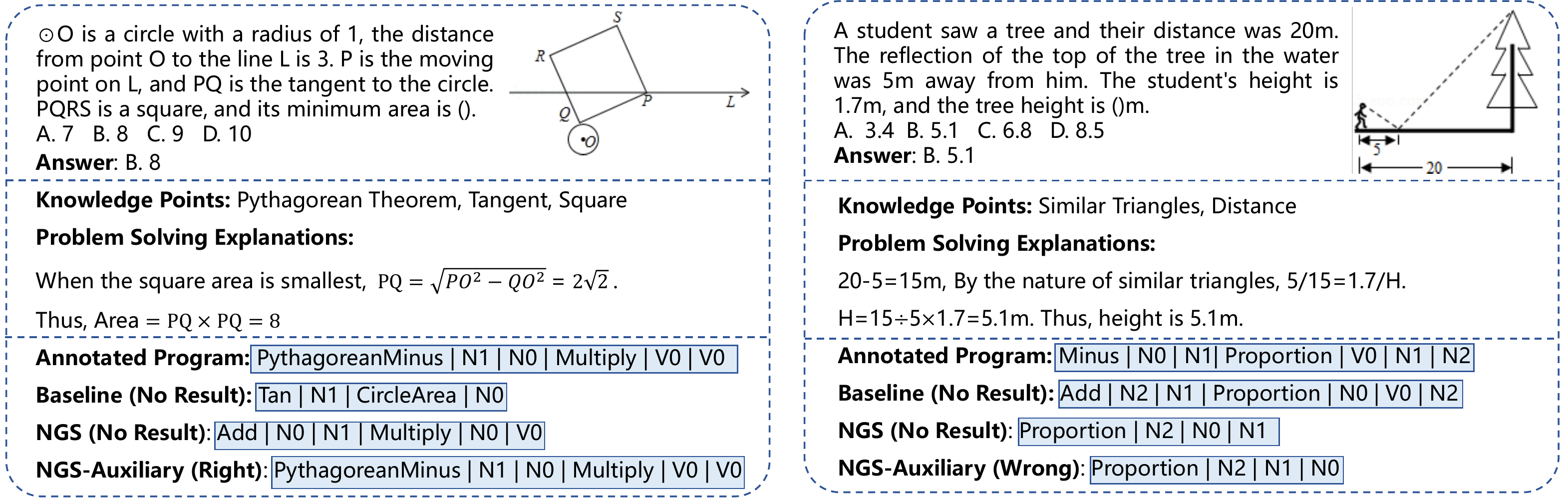}
\end{center}
  \caption{Typical cases. No Result represents the answer executed by the programs is not in the options, and Wrong represents getting a wrong option. Baseline is a ``Seq2Prog + Diagram" model. In the case on the left, NGS-Auxiliary successfully predicts the knowledge of the Pythagorean theorem through auxiliary tasks and get the right answer. For the case on the right, the problem is quite hard that current model cannot solve it. 
  }
\label{fig:case}
\end{figure*}

\begin{figure}[t]
\begin{center}
 \includegraphics[width=0.9\columnwidth]{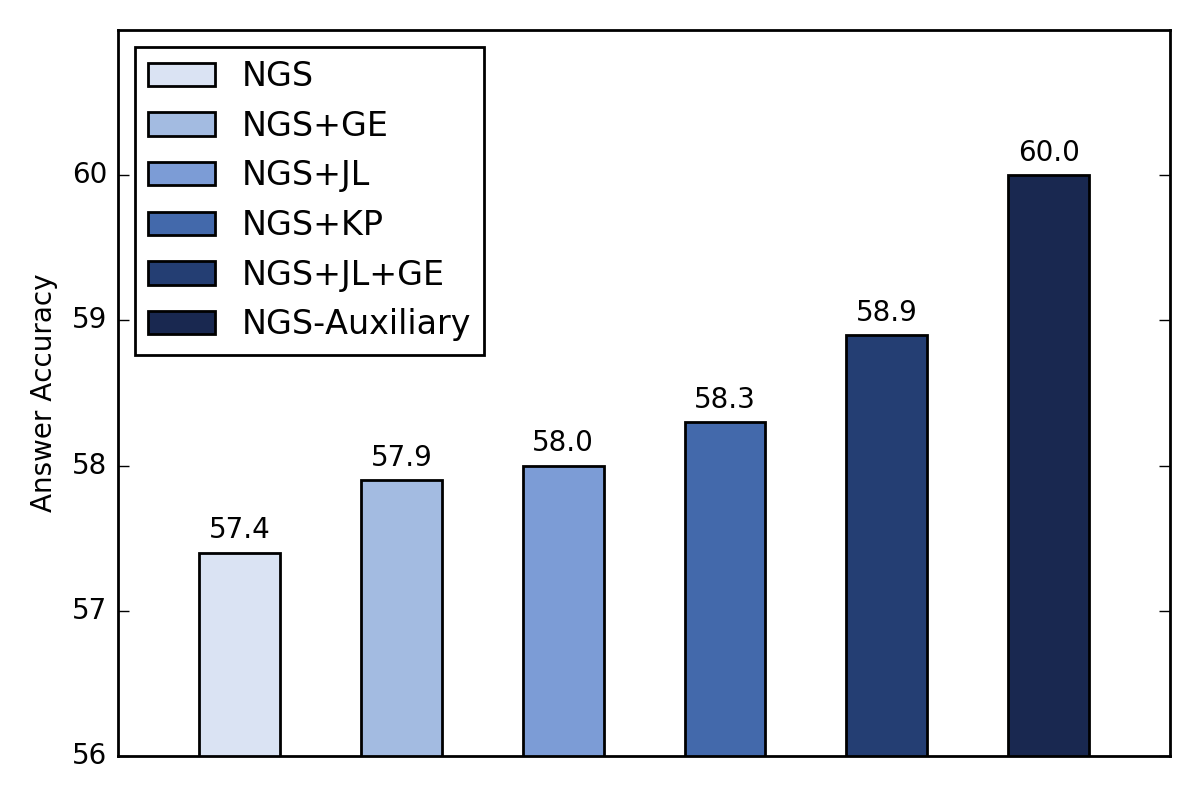}
\end{center}
  \caption{Ablation study on different auxiliary
components. ‘+’ represents we add the auxiliary component. NGS-Auxiliary means that adding all three auxiliary tasks together.
  }
\label{fig:ablation}
\end{figure}

\subsection{Experimental Results}

We introduce three types of models here and test them on our GeoQA.
The performance comparison with various methods on the different subset types of GeoQA is reported in Table \ref{table-baselines}. 

\textbf{Human Performance.}
We invite 10 students with a high score (top 1\%) in the national university entry exam to answer these geometric problems. 
For each question, they first try to solve the problem with the text only and draw a diagram by themselves. Then, the actual diagram is given to answer the complete question. The total time for each question is limited to two minutes.  When using text and diagram simultaneously, the human performance is improved from 63.0\% to 92.3\%, which indicates that humans are not good at solving problems using only text, but handling multimodal information successfully.
The result shows that there is still a huge gap between the existing models and human experts, leaving large room for future research.

\textbf{The effectiveness of program.} ``W/O Program" refers to not using programs and regarding GeoQA as a classification problem similar to VQA. Therefore, we conduct experiments on three models with multimodal reasoning capabilities, including FiLM, RN, and MCAN. However, their performance results show that these methods fail to reason about such complex geometric problems, achieving poor performance on GeoQA. 
These results prove the effectiveness and importance of our designed interpretable programs.

\textbf{The necessity of multi-modality.} ``Text-Only" means that models only use text to generate program sequences since humans can also understand the intent of the question, draw diagram based on the text, and solve the problem. Motivated by~\citet{amini2019mathqa}, we design a Sequence-to-Program (Seq2Prog) model using GRU encoder with an attention mechanism. Moreover, we replace the encoder with BERT and get a stronger BERT2Prog. By predicting our tailor-designed programs, these two methods are effective, while the performance is not satisfactory enough. 
These results show that multimodal reasoning ability is indispensable when solving geometric problems.

%


\textbf{Multimodal numerical reasoning baselines. }``Text-Diagram" refers to using text and diagram simultaneously. We concatenate the text embedding with the diagram feature extracted by ResNet \cite{he2016deep} as the baseline methods, such as Seq2Prog + Diagram and BERT2Prog + Diagram. 
These feature fusion methods do not have the strong reasoning ability and fail to improve the program decoding. 
For instance, by adding diagram, the performance of BERT2Prog + Diagram declines from 54.4\% to 52.5\%, which may result from the extra diagram that disturbs the text pretraining model.
Note that multimodal pretraining models~\cite{lu2019vilbert,li2020unicoder} cannot be applied to geometric problems, since these models are based on Faster-RCNN to extract object-level features from natural images.

\textbf{The effectiveness of our methods.}
Our proposed NGS shows a relatively-good performance compared to the various models mentioned above. When adding all three auxiliary tasks together to enhance NGS solver, our NGS-Auxiliary with multimodal reasoning ability  becomes the existing best-performing method (60.0\%) on GeoQA dataset.
It also achieves the highest accuracy on all types of problems. For example, compared with Seq2Prog+Diagram, NGS-Auxiliary obtains an 9.1\% performance improvement on the angle type problems.
Compared with other ``Text-Diagram" baselines, our model is effective when reasoning on multimodal information.

\textbf{The effect of different beam size. }
In general, we set the beam size to 10 for testing. 
In this section, we explore the influence of different beam size.
After the searched sequence program is executed, there will be three situations: right answer, wrong answer, and no result. 
As shown in Table.~\ref{table-beam}, as the beam size is larger, we get higher accuracy and a lower proportion of no result. When beam size equals 1, the  NGS-Auxiliary outperforms baselines significantly. Our model can achieve the highest accuracy of 63.9\%  when beam size is 100. 

\subsection{Ablation Study}
As shown in Fig.~\ref{fig:ablation}, we conduct experiments to evaluate the contribution of different auxiliary tasks. We consider six different combinations: 1) only the NGS; 2) NGS + Geometry Elements (NGS+GE); 3) NGS + Jigsaw Location (NGS+JL);  4) NGS + Knowledge Points (NGS+KP); 5) NGS + diagram-based pretraining (NGS+JL+GE); 6) NGS with all three auxiliary tasks (NGS-Auxiliary).
We can see that all three auxiliary tasks can promote the performance of NGS. The accuracy gains of GE, JL, KP, JL + GE, and combining all three tasks are 0.5\%, 0.6\%, 0.9\%, 1.5\%, 2.6\%, respectively. 
These results show that all our self-supervised and auxiliary tasks can enhance the comprehensive understanding and multimodal reasoning ability of NGS. 

\subsection{Case Analysis}
As shown in Fig.~\ref{fig:case}, we select two typical cases to demonstrate the programs generated by different models and some representative errors. 

In the left case, the knowledge of the Pythagorean theorem, tangent, and square are required for solving the problem. 
Baseline method and our NGS fail to generate the correct operations and get no result. 
However, our NGS-Auxiliary successfully predicts the use of knowledge in the proposed auxiliary task. And more importantly, the correct "PythagoreanMinus" program is generated, and the right answer is obtained. 

The right one is a typical error case, in which model needs to understand a  complex scene. Although NGS-Auxiliary has predicted the knowledge points of Proportion program correctly, all three models fail to predict the correct answer. A better multimodal method is required to handle this hard high-level reasoning task in the future.

\section{Conclusion}
In this work, we focus on the geometric problem and propose the first large-scale geometric question answering dataset ``GeoQA", containing 4,998 problems with program annotation. Besides, we propose a deep neural baseline, named as Neural Geometric Solver (NGS), to solve a geometric problem by jointly reasoning over multimodal data and generating interpretable programs. We further propose multiple novel auxiliary tasks to enhance the semantic representation of text and diagram. 
Extensive experimental results and analyses show that our GeoQA is challenging, and our NGS-Auxiliary outperforms other methods on GeoQA.

\section{Ethical Impact}
We collected GeoQA from two online education websites, which is only used for academic research, and the copyright belongs to the original websites. This work may inspire research in the field of multimodal numerical reasoning.

\paragraph{Acknowledgements}
This work was supported in part by National Key R\&D Program of China under Grant No.2020AAA0109700, National Natural Science Foundation of China (NSFC) under Grant No.U19A2073 and No.61976233, Guangdong Province Basic and Applied Basic Research (Regional Joint Fund-Key) Grant No.2019B1515120039,  Shenzhen Fundamental Research Program (Project No.RCYX20200714114642083 and No.JCYJ20190807154211365), Zhijiang Lab’s Open Fund (No.2020AA3AB14), and CSIG Young Fellow Support Fund.


\bibliographystyle{acl_natbib}
\bibliography{acl2021}
\end{document}